\documentclass[runningheads]{llncs}
\usepackage{amsmath}
\usepackage{amssymb}
\usepackage{bbm}
\usepackage{graphicx}
\usepackage{algpseudocode}
\usepackage{soul}
\usepackage{multirow}
\usepackage{authblk}

\begin{document}
\title{C$^3$PS: Context-aware Conditional Cross Pseudo Supervision for Semi-supervised Medical Image Segmentation}
%
%
\author{Peng Liu}
\author{Peng Liu and Guoyan Zheng\thanks{Corresponding Author: guoyan.zheng@sjtu.edu.cn}}

\authorrunning{Liu. Author et al.}
%
\institute{Shanghai Jiao Tong University}

\maketitle              
\begin{abstract}
Semi-supervised learning (SSL) methods, which can leverage a large amount of unlabeled data for improved performance, has attracted increasing attention recently. In this paper, we introduce a novel Context-aware Conditional Cross Pseudo Supervision method (referred as C$^3$PS) for semi-supervised medical image segmentation. Unlike previously published Cross Pseudo Supervision (CPS) works, this paper introduces a novel Conditional Cross Pseudo Supervision (CCPS) mechanism where the cross pseudo supervision is conditioned on a given class label. Context-awareness is further introduced in the CCPS to improve the quality of pseudo-labels for cross pseudo supervision. The proposed method has the additional advantage that in the later training stage, it can focus on the learning of hard organs. Validated on two typical yet challenging medical image segmentation tasks, our method demonstrates superior performance over the state-of-the-art methods.

\end{abstract}
\section{Introduction}
Medical Image Segmentation (MIS) is an essential step in many clinical applications. Past years witnessed remarkable progress of application of supervised deep learning (DL)-based methods to medical image segmentation. However, supervised DL-based methods typically require a large amount of expert-level accurate, densely-annotated data for training, which is laborious and costly to collect. To this end, various annotation-efficient methods have been introduced \cite{yao2021label,zhang2020generalizing,zhang2021exploiting}. Among them, Semi-Supervised MIS (SSMIS) methods have attracted increasing attention as they only require a limited number of labeled data while leveraging a large amount of unlabeled data for improved performance. As a SSL-based image segmentation method, Cross Pseudo Supervision (CPS) \cite{chen2021semi} has set the new state-of-the-art (SOTA) in the task of semi-supervised semantic segmentation of natural images \cite{xu2021learning}. CPS works by imposing the consistency on two segmentation networks perturbed with different initialization for the same input image. The pseudo one-hot label map, generated from one perturbed segmentation network, is used to supervise the training of the other segmentation network, and vice versa. The idea behind the CPS consistency is that it can encourage high similarity between the predictions of two perturbed networks for the same input image while expanding training data by using the unlabeled data with pseudo labels. 

Inspired by the original idea introduced by Chen et al. \cite{chen2021semi}, various CPS-based approaches have been proposed in the literature for SSMIS \cite{luo2021semi,LinYLZL22,liu2022semi}. For example, Luo et al.\cite{luo2021semi} introduced cross teaching between a Convolutional Neural Network (CNN) and a Transformer for SSMIS. Lin et al. \cite{LinYLZL22} proposed a framework based on CPS which used class-aware weighted loss, probability-aware random cropping, and dual uncertainty-aware sampling supervision to conduct semi-supervised segmentation of knee joint MR images. Liu et al.\cite{liu2022semi} proposed a framework based on CPS that generated anatomically plausible predictions using shape awareness and local context constraints. Despite these progress, however, there are still space for further improvement. Specifically, how to design better cross pseudo supervision strategy is still an open problem.

In this paper, we propose a Context-aware Conditional Cross Pseudo Supervision method, referred as C$^3$PS, for SSMIS. Unlike previous works \cite{chen2021semi,luo2021semi,LinYLZL22,liu2022semi}, we propose a novel Conditional Cross Pseudo Supervision (CCPS) mechanism where the cross pseudo supervision is conditioned on a given class label. Inspired by \cite{lai2021semi}, context-awareness is further introduced in the CCPS to improve the quality of pseudo-labels for cross pseudo supervision. The proposed method has the additional advantage that in the later training stage, it can focus on the learning of hard organs. Our contributions can be summarized as follows:

\begin{itemize}
	\item We propose a context-aware conditional cross pseudo supervision method, referred as C$^3$PS, for semi-supervised medical image segmentation. C$^3$PS is based on cross teaching between a regular CNN (RNet) and a conditional CNN (CNet) where RNet is a multi-class segmentation network while CNet is a binary segmentation network conditioned on a given class label. 
	\item Based on the network design, we further introduce a novel CCPS mechanism where the cross pseudo supervision is conditioned on a given class label. CCPS mechanism has the additional advantage that in the later training stage, it can focus on the learning of hard organs. Context-awareness is further introduced in the CCPS to improve the quality of pseudo-label generation.
	\item We validate C$^3$PS on two typical yet challenging SSMIS tasks.
\end{itemize}

\section{Method}

\begin{figure*}[tbp]
  \centering
  \includegraphics[width=0.9\textwidth]{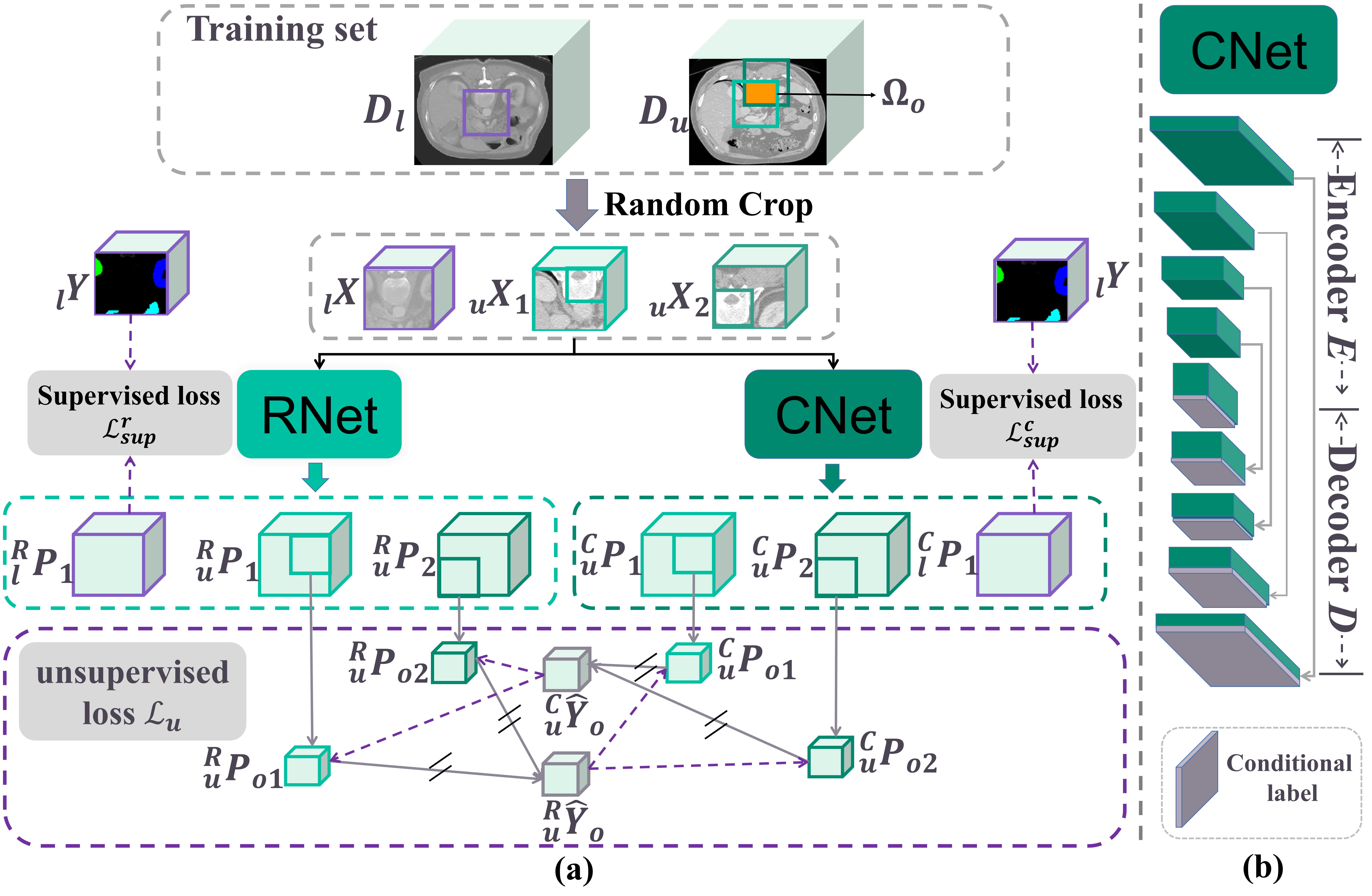}
  \caption{Overview of the proposed C$^3$PS framework. (a) C$^3$PS framework. (b) architecure of CNet. When inputting to the CNet, we incorporate conditional label in the decoder path. `$//$' on $\rightarrow$ means combine two predictions into one prediction and stop-gradient. `$\dashrightarrow$' means loss supervision.} \label{method_overall}
\end{figure*}

As illustrated in Fig. \ref{method_overall}-(a), C$^3$PS consists of two networks: RNet and CNet. CNet is a binary segmentation network, which generates segmentation for a given conditional class label. We leverage unlabeled data by implementing CCPS between these two networks. Context-awareness is further introduced in the CCPS to improve the quality of pseudo-label generation. Formally, we denote RNet as $\mathcal{F}^R(.)$ with weights $\Theta_R$, CNet as $\mathcal{F}^C(.)$ with weights $\Theta_C$.
Denote $\mathcal{D}_{l}$ the labeled dataset, 
$\mathcal{D}_{u}$ the unlabeled dataset. We train our model with $\mathcal{D}_{l}$ and $\mathcal{D}_{u}$. 

\subsection{Supervised learning}
We sample a labeled patch $(_lX,_lY)$ from $\mathcal{D}_l$. Given a conditional label $c_l$ (details on generating $c_l$ will be described below). We can get predictions from $\mathcal{F}^R(.)$ and $\mathcal{F}^C(.)$ as follows:
\begin{equation}
    ^R_l\mathbf{P}=\mathcal{F}^R(_lX;\Theta_R); \quad
    ^C_l\mathbf{P}=\mathcal{F}^C(_lX,c_l;\Theta_C)
\end{equation}
 The supervised loss $\mathcal{L}^r_{sup}$ for RNet and $\mathcal{L}^c_{sup}$ for CNet can be defined respectively as follows:
\begin{equation}
    \mathcal{L}^R_{sup} = \mathcal{L}_{ce}(^R_l\mathbf{P}, {_lY}) + \mathcal{L}_{dice}(^R_l\mathbf{P}, {_lY})
\end{equation}

\begin{equation}
     \mathcal{L}^C_{sup} = \mathcal{L}_{ce}(^C_l\mathbf{P}, \mathbbm{1}\{_lY = c_l\}) + \mathcal{L}_{dice}(^C_l\mathbf{P}, \mathbbm{1}\{_lY = c_l\})
\end{equation}
where $\mathcal{L}_{ce}$ and $\mathcal{L}_{dice}$ denote respectively the cross entropy loss and the Dice loss. $\mathbbm{1}\{_lY = c_l\}$ aims to generate a binary mask depending on whether the label of a voxel is equal to $c_l$ or not. Finally, the overall supervised loss $\mathcal{L}_{sup}$ can be defined as follows:
\begin{equation}
    \mathcal{L}_{sup} = \mathcal{L}^R_{sup} + \mathcal{L}^C_{sup}
\end{equation}

\subsection{Conditional Cross Pseudo Supervision}
 \subsubsection{CCPS loss between two networks.}
We sample two unlabeled patches $_uX_1$ and $_uX_2$ from $\mathcal{D}_u$, where $_uX_i \in \mathbb{R}^{H \times W \times D},i=1,2$. Given a conditional label $c_u$ (details on generating $c_u$ will be described below), we can get predictions for $_uX_i$ from $\mathcal{F}^R(.)$ and $\mathcal{F}^C(.)$ as follows:
 \begin{equation}
 \label{Eq_5}
     ^R_u\mathbf{P}_i=\mathcal{F}^R(_uX_i;\Theta_R); \quad ^C_u\mathbf{P}_i=\mathcal{F}^C(_uX_i,c_u;\Theta_C);
 \end{equation}

We further obtain the pseudo labels generated by these two networks as follows:
 \begin{equation}
     ^R_u\hat{Y}_i = argmax(^R_u\mathbf{P}_i); \quad ^C_u\hat{Y}_i = argmax(^C_u\mathbf{P}_i);
 \end{equation}
 where $^C_u\hat{Y}_i$ is a binary segmentation mask. 

We then conduct cross pseudo supervision between the RNet and the CNet. Concretely, we first use $^C_u\hat{Y}_i$ as pseudo labels to supervise the learning of the RNet:
\begin{equation}
    \mathcal{L}^R_{ce}({^R_u\mathbf{P}_i}, {^C_u\hat{Y}_i}, {^C_u\mathbf{P}_i}) = \sum^{H \times W \times D}_{j=1} {\mathbbm{1}\{^C_u\mathbf{p}^1_{i,j} > \gamma_1\}{^C_u\hat{Y}_{i,j}} log ({^R_u\mathbf{p}^{c_u}_{i,j}})}
\end{equation}
where $^C_u\mathbf{p}^1_{i,j}$ means the predicted probability from the CNet for the conditional class $c_u$ at the $j$-th voxel of image $i$; $\mathbbm{1}\{^C_u\mathbf{p}^1_{i,j} > \gamma_1\}$ is used to ensure the quality of the pseudo label $^C_u\hat{Y}_{i,j}$ at the $j$-th voxel of image $i$; $\gamma_1$ is a confidence threshold; $^R_u\mathbf{p}^{c_u}_{i,j}$ means the predicted probability from the RNet for the conditional class label $c_u$ at the $j$-th voxel of image $i$. Please note that when we use the output from the CNet to supervise the learning of the RNet, we only compute loss for foreground class since the predicted background class of the CNet may contain other organs.


Then we use $^R_u\hat{Y}_i$ as pseudo labels to supervise the learning of the CNet:
\begin{equation}
\begin{split}
    \mathcal{L}^C_{ce}({^C_u\mathbf{P}_i}, {^R_u\hat{Y}_i}, {^R_u\mathbf{P}_i}) =  \sum^{H \times W \times D}_{j=1} (&\mathbbm{1}\{^R_u\mathbf{p}^{c_u}_{i,j} > \gamma_2\}  \cdot{\mathbbm{1}\{^R_u\hat{Y}_{i,j}=c_u\} log (^C_u\mathbf{p}_{i,j}^1}) + \\ &\mathbbm{1}\{1-^R_u\mathbf{p}^{c_u}_{i,j} > \gamma_2\} \cdot {\mathbbm{1}\{^R_u\hat{Y}_{i,j} \neq c_u\} log (^C_u\mathbf{p}_{i,j}^0}))
\end{split}
\end{equation}
where $^R_u\mathbf{p}^{c_u}_{i,j}$ means the predicted probability from the RNet for the conditional class $c_u$ at the $j$-th voxel of image $i$; $\mathbbm{1}\{^R_u\mathbf{p}^{c_u}_{i,j} > \gamma_2\}$ and $\mathbbm{1}\{(1-^R_u\mathbf{p}^{c_u}_{i,j}) > \gamma_2\}$ are used to ensure the quality of the pseudo label $^R_u\hat{Y}_{i,j}$ at the $j$-th voxel of image $i$; $\gamma_2$ is a confidence threshold; $^C_u\mathbf{p}_{i,j}^1$ and $^C_u\mathbf{p}_{i,j}^0$ denote the predicted probabilities from the CNet for the conditional class $c_u$ and other classes at the $j$-th voxel of image $i$, respectively.

Then our CCPS loss $\mathcal{L}_{ccps}$ for the unlabeled data is defined as:
\begin{equation}
    \mathcal{L}_{ccps} = \frac{1}{2}\sum^{2}_{i=1}\mathcal{L}^R_{ce}({^R_u\mathbf{P}_i}, {^C_u\hat{Y}_i}, {^C_u\mathbf{P}_i}) +  \\
		\frac{1}{2}\sum^{2}_{i=1}\mathcal{L}^C_{ce}({^C_u\mathbf{P}_i}, {^R_u\hat{Y}_i}, {^R_u\mathbf{P}_i})
\end{equation}
 \subsubsection{Strategy for generating conditional labels.} Since we sample patches for model training, each sampled patch may only contain a subset of overall label set $\{1,2,3,...,L\}$. In order to make our training process more efficient, we design a strategy to generate conditional labels for a labeled patch $(_lX,{_lY})$ and an unlabeled patch $_uX$. For the labeled patch $(_lX,{_lY})$, we first  get unique label set from $_lY$: $\mathbb{S}_l=unique(_lY)$, where $unique( \cdot )$ returns the unique values in the label set. Then we randomly select $c_l$ from $\mathbb{S}_l$. For the unlabeled patch $_uX$, we first get unique label set from $^R_u\hat{Y}$: $\mathbb{S}_u=unique(^R_u\hat{Y})$. Then we randomly select $c_u$ from $\mathbb{S}_u$.
 
\subsubsection{Hard Organ Learning (HOL) with CCPS.}
Since the CNet can learn with a given conditional label, in the later training stage when both the RNet and the CNet can generate stable segmentation results, we can feed a label subset to the CNet and let the CNet focus on the classes which are hard to segmentation (e.g., in abdominal organ segmentation, pancreas and kidneys are deemed to be more difficult due to their relatively small sizes). Concretely, we first define the label subset $\mathbb{S}_{s}$ for hard organs. Then, for labeled data $_lX$, we randomly select a conditional label $c_l$ from label set $\mathbb{S}_l \bigcap \mathbb{S}_{s}$. Similarly, for unlabeled data $x_u$, we randomly select a conditional label $c_u$ from label set $\mathbb{S}_u \bigcap \mathbb{S}_{s}$. We set $\mathbb{S}_{s}$ as the full label set in the first $t_s$ iterations. After that, we set $\mathbb{S}_{s}$ as the hard organ label subset.


\subsection{Context-awareness for pseudo-label generation}
When patch-based strategy is used to train a network, predictions of a voxel may be different when the context is different (e.g., in different patches). Liu et al. \cite{liu2022context} sampled two overlapped patches in each training iteration to make the model more robust to the context change. Followed by this work, we incorporate context-awareness to increase the quality of pseudo label generation. Specifically, for each iteration, we require that two sampled patches $_uX_1$ and $_uX_2$ have an overlapping region $\Omega_o$ (shown in $\mathcal{D}_u$ of Fig. \ref{method_overall}-(a)). We further require that the CCPS loss defined above is only computed in the region $\Omega_o$. Then we feed $_uX_1$ and $_uX_2$ to $\mathcal{F}^R(.)$ and $\mathcal{F}^C(.)$ and get predictions using Eq. (\ref{Eq_5}):

The predictions of the overlapped region in $_uX_1$ and $_uX_2$ from the RNet are $^R_u\mathbf{P}_{o1}$ and $^R_u\mathbf{P}_{o2}$, respectively. Similarly, the predictions of the overlapped region in $_uX_1$ and $_uX_2$ from the CNet are $^C_u\mathbf{P}_{o1}$ and $^C_u\mathbf{P}_{o2}$, respectively. We generate pseudo labels of voxels in $\Omega_o$ by combining two predictions with different context as follows:
\begin{equation}
^R_u\mathbf{P}_{o} = (^R_u\mathbf{P}_{o1} + ^R_u\mathbf{P}_{o2})/2; \quad
    ^R_u\hat{Y}_o = argmax(^R_u\mathbf{P}_{o})
\end{equation}
\begin{equation}
^C_u\mathbf{P}_{o} = (^C_u\mathbf{P}_{o1} + ^C_u\mathbf{P}_{o2})/2; \quad
    ^C_u\hat{Y}_o = argmax(^C_u\mathbf{P}_{o})
\end{equation}

Finally, the loss $\mathcal{L}_{C^3PS}$ for the unlabeled data after incorporating context-awareness is computed as follows:
\begin{equation}
\begin{split}
    \mathcal{L}_{C^3PS} = \frac{1}{2}(&\mathcal{L}^R_{ce}({^R_u\mathbf{P}_{o1}}, {^C_u\hat{Y}_o}, {^C_u\mathbf{P}_{o}}) + \mathcal{L}^R_{ce}({^R_u\mathbf{P}_{o2}}, {^C_u\hat{Y}_o}, {^C_u\mathbf{P}_{o}}) + \\
    &\mathcal{L}^C_{ce}({^C_u\mathbf{P}_{o1}}, {^R_u\hat{Y}_o}, {^R_u\mathbf{P}_{o}}) + \mathcal{L}^C_{ce}({^C_u\mathbf{P}_{o2}}, {^R_u\hat{Y}_o}, {^R_u\mathbf{P}_{o}}))
\end{split}
\end{equation}

\subsection{Overall loss function}
The overall training objective $\mathcal{L}$ of our proposed approach is:
\begin{equation}
    \mathcal{L} = \mathcal{L}_{sup} + \lambda\mathcal{L}_{C^3PS}
\end{equation}
where $\lambda$ is a weight factor defined as: $\lambda(t) = 0.1 \times e^{(-5(1-\frac{t}{t_{total}})^2)} $, where $t$ denotes the current iteration and $t_{total}$ is the total iteration number.

\begin{table}[t]
  \caption{\label{overall_results} Comparisons with the SOTA methods on the BCV and the MMWHS datasets. L: number of labeled data, U: number of unlabeled data.}
  \centering
\begin{tabular}{c|c|c|c|c|c|c|c|c}
\hline
\multirow{2}*{Method} & \multicolumn{4}{c|}{BCV}        & \multicolumn{4}{c}{MMWHS}                 \\  \cline{2-9}
    &L &U & \multicolumn{1}{c|}{DSC (\%)$\uparrow$}  & \multicolumn{1}{c|}{ASD (voxel)$\downarrow$} &L &U      & DSC (\%)$\uparrow$  &ASD (voxel)$\downarrow$             \\ \hline
\multicolumn{1}{c|}{Baseline} &4 &0   &63.7     &11.19 &2 &0   &75.6 & 28.66            \\ 
Upper Bound &24 &0   &86.9       & 10.31 &14 &0     &91.2  & 4.13         \\ \hline
DAN \cite{zhang2017deep}&4 &20  &64.2 &11.92 &2 &12 &75.4 & 37.50  \\
MT \cite{tarvainen2017mean}&4 &20 &65.2   &11.15 &2 &12   &78.1   &28.26  \\
UAMT \cite{yu2019uncertainty}&4 &20  &67.6  &13.59 &2 &12  &77.7  &29.01    \\
URPC \cite{luo2021efficient}&4 &20   &65.7 &15.08 &2 &12 &72.9 &36.66   \\
McNet \cite{wu2022mutual}&4 &20   &60.4      &10.49 &2 &12    &67.7   &21.38  \\
\multicolumn{1}{c|}{CPS \cite{chen2021semi}} &4 &20 &67.2  &11.45 &2 &12    &78.9 &27.23  \\
\multicolumn{1}{c|}{Ours}   &4 &20 &\textbf{72.4} &\textbf{9.18} &2 &12 &\textbf{81.7} &\textbf{11.06} \\ \hline
\end{tabular}
\end{table}

\begin{table}[tb]
  \caption{\label{BCV_different_components} Results of investigating the effectivensss of different components. }
  \centering
\begin{tabular}{c|c|c|c|cc}
\hline
CPS & CCPS & Context-awareness & HOL & DSC (\%)$\uparrow$ & ASD (voxel)$\downarrow$ \\ \hline
\checkmark  &             &                &             & 67.2    &11.45   \\
\checkmark  & \checkmark  &                &             &67.8     &9.75    \\
\checkmark  &             &\checkmark      &             &68.5     &9.80   \\
\checkmark  &\checkmark   &\checkmark      &             &71.6     &10.91  \\
\checkmark  &\checkmark   &\checkmark      &\checkmark   &\textbf{72.4}     &\textbf{9.18}   \\ \hline
\end{tabular}
\end{table}

\section{Experiments}
\subsection{Dataset and Implementation details}
\textbf{Datasets.} 
We evaluated our method on two public datasets: Beyond the Cranial Vault (BCV) dataset\cite{gibson2018automatic}  and The MICCAI'17 Multi-Modality Whole Heart Segmentation challenge (MMWHS) dataset\cite{zhuang2016multi}. The BCV dataset contains 30 CT images with annotation for 13 organs. We chose to segment five abdominal organs including liver, spleen, pancreas, left kidney and right kidney. We used 24 samples (4 labeled and 20 unlabeled) for training and the remaining 6 samples for testing. The MMWHS consists of 20 cardiac CT samples with annotations for seven structures: left ventricle (LV), right ventricle (RV), left atrium (LA), right atrium (RA), pulmonary artery (PA), my-ocardium (MYO) and ascending aorta (AA). We took 14 samples (2 labeled and 12 unlabeled) for training and used the remaining 6 samples for testing.

\textbf{Implementation details.} 
We chose 3D U-Net\cite{ronneberger2015u} as the RNet and the conditional 3D U-Net (see Fig. \ref{method_overall}-(b) for an illustration) as the CNet \cite{zhang2021multiorgan}. Please note, for a fair comparison with other SOTA SSMIS methods, we replaced the nnUNet used in \cite{zhang2021multiorgan} by 3D U-Net. We used a patch size of 160 $\times$ 160 $\times$96 for training. In total we trained our network 20000 iterations. We used SGD optimizer and set the initial learning rate to 0.01. Then at each iteration the initial learning rate was multiplied by $(1.0-\frac{t}{20000})^{0.9}$, where $t$ is current iteration. We implemented our method based on PyTorch framework and conducted the evaluation on a NVIDIA Tesla V100 GPU. Starting from ($t_s=15000$)th iteration, we conducted hard organ learning. On the BCV dataset, we chose left kidney, right kidney and pancreas as the hard organs while on the MMWHS dataset, we chose MYO, RA, AA and RV as the hard organs. We empirically set the thresholds $\gamma_1$ and $\gamma_2$ to 0.95 and 0.9, respectively. We use Dice score coefficient (DSC; \%) and Average Surface Distance (ASD; Voxel) as the evaluation metrics. Paired t-test is used to check whether a difference is statistically significant or not. We set the significant level as 0.05.

\begin{figure*}[t] 
  \centering
  \includegraphics[width=\textwidth]{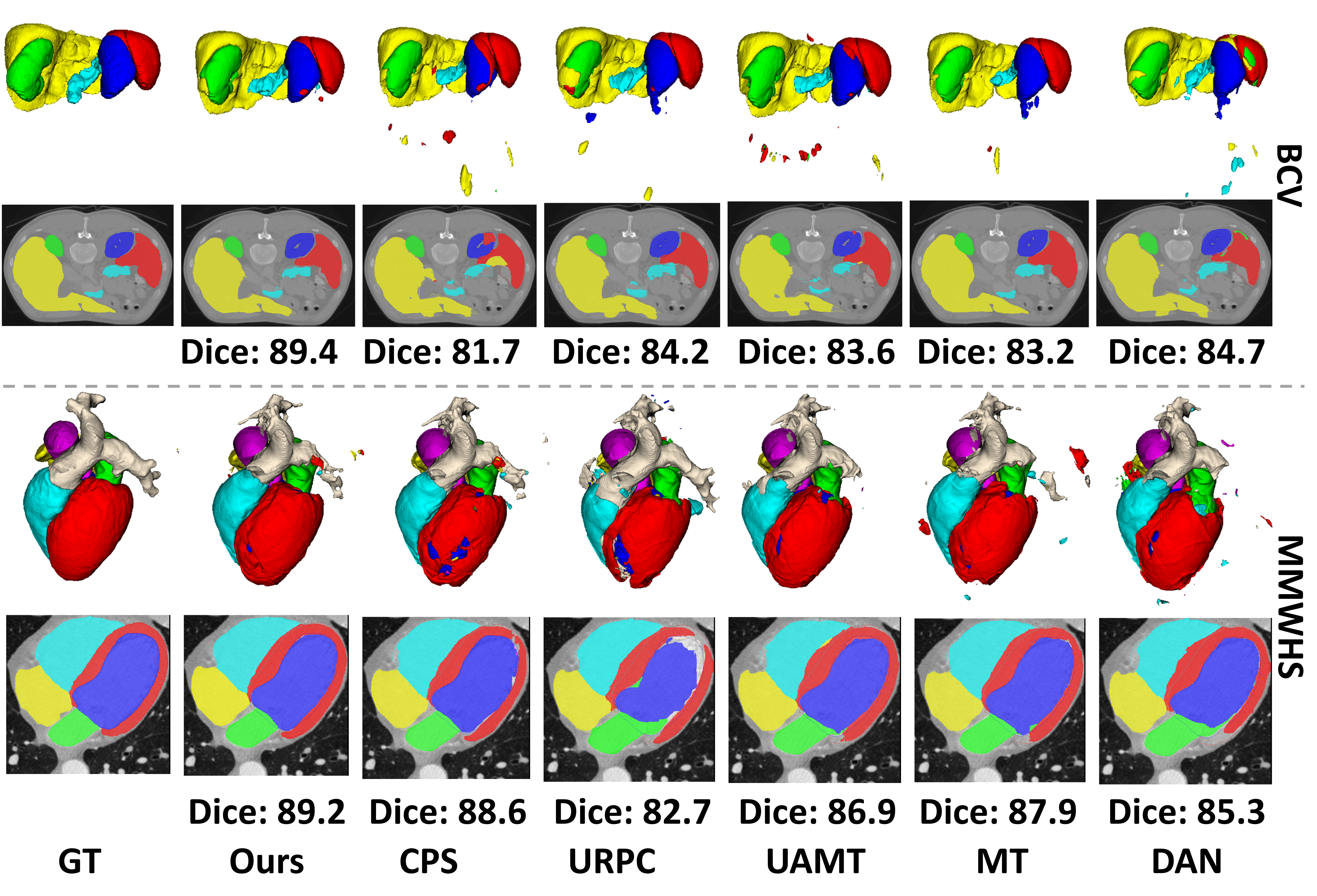}
  \caption{Visual comparison of different SSL-based segmentation methods on both the BCV and the MMWHS datasets.} \label{BCV_visualization}
\end{figure*}

\subsection{Results}
We compared our methods with six SOTA SSMIS methods \cite{zhang2017deep,tarvainen2017mean,yu2019uncertainty,luo2021efficient,chen2021semi,wu2022mutual}. Baseline and Upper Bound are obtained by training the RNet in a supervised manner. 

\textbf{BCV dataset.} 
Table \ref{overall_results} presents the results on the BCV dataset. It can be found that the proposed C$^3$PS method achieves better DSC performance than other 6 SOTA methods with a large margin when 4 labeled data were used. Additionally, from this table, one can also find that our method outperforms CPS with a large margin in terms of DSC (on average an increase of 5.2\%). Paired t-test showed that the difference between our method and the CPS method \cite{chen2021semi} was statistically significant ($p$=0.011). 

\textbf{MMWHS dataset.} 
Table \ref{overall_results} also presents the results on the MMWHS dataset. The proposed method achieved a superior performance over other SOTA methods with an average DSC of 81.7\%. Paired t-test showed that the difference between our method and the CPS method \cite{chen2021semi} was statistically significant ($p$=1e-4). 

Fig. \ref{BCV_visualization} shows a visual comparison of the top-5 methods as well as our own method when applied to these two datasets. From this figure, one can see that our method predicts more consistent segmentation results with the ground truth than other methods.

\textbf{Ablation Study.} 
We conducted ablation studies on the BCV dataset to show the effectiveness of each individual components. As shown in Table \ref{BCV_different_components}, each component contributes to the superior performance of the proposed approach. We further investigate the effectiveness of HOL. The results are presented in Table \ref{BCV_dice_of_different_method}. With HOL, we observed an improved performance for left and right kidney as well as pancreas. 

\begin{table}[t]
 \caption{\label{BCV_dice_of_different_method} Results of investigating the effectivenss of hard organ learning. w/o: without; w/: with}
 \centering
\begin{tabular}{c|ccccc|c}
\hline
Method                       & Spleen        & Right Kidney      & Left Kidney   & Liver     & Pancreas    & All  \\ \hline
Ours (w/o HOL)   & \textbf{79.9}      &  78.2     & 73.4  & 82.3    & 44.3         & 71.6 \\
Ours (w/ HOL)     & 78.8   & \textbf{78.6}   & \textbf{74.9} & \textbf{83.5}   & \textbf{46.3}  & \textbf{72.4}  \\
\hline
\end{tabular}
\end{table}

\section{Conclusion}
In this paper, we proposed a context-aware conditional cross pseudo supervision method, referred as C$^3$PS, for semi-supervised medical image segmentation. We further introduced a novel CCPS mechanism where the cross pseudo supervision was conditioned on a given class label. CCPS mechanism had the additional advantage that in the later training stage, it could focus on the learning of hard organs. Context-awareness was further introduced in the CCPS to improve the quality of pseudo-label generation. Results from experiments conducted on two public datasets demonstrate the efficacy of the proposed approach.

\bibliographystyle{unsrt}
\bibliography{References}
\end{document}